\documentclass[
]{ceurart}

\usepackage{listings}
\usepackage{amsmath}
\usepackage{amssymb}
\usepackage{tipa}
\usepackage{tcolorbox}

\begin{document}

\copyrightyear{2026}
\copyrightclause{Copyright for this paper by its authors.
  Use permitted under Creative Commons License Attribution 4.0
  International (CC BY 4.0).}

\conference{EVALITA 2026: 9th Evaluation Campaign of Natural Language
Processing and Speech Tools for Italian, Feb 26 – 27, Bari, IT}

\title{AIWizards at MULTIPRIDE: A Hierarchical Approach to Slur Reclamation Detection}

\author[1,3]{Luca Tedeschini}[%
orcid=0009-0006-0375-829X,
email=ltedeschini@villanova.ai,
url=https://github.com/LucaTedeschini,
]
\cormark[1]
\fnmark[1]

\author[2]{Matteo Fasulo}[%
orcid=0000-0002-7019-3157,
email=mfasulo@ethz.ch,
url=https://www.matteofasulo.com,
]
\address[1]{Villanova.ai S.P.A, Località Sa Illetta, SS 195 KM 2.3, 09123 Cagliari, Italy}
\address[2]{Swiss Data Science Center, ETH Z{\"u}rich, Andreasturm, Andreasstrasse 5, 8092 Z{\"u}rich, Switzerland}
\address[3]{Department of Computer Science and Engineering (DISI), University of Bologna, Mura Anteo Zamboni 7, 40126 Bologna, Italy}
\fnmark[1]

\cortext[1]{Corresponding author.}
\fntext[1]{These authors contributed equally.}

\begin{keywords}
    Hate Speech Detection \sep
    Reclaimed Slurs \sep
    User Profiling \sep
    Hierarchical Modeling \sep
    Social Media NLP
\end{keywords}

\maketitle

\begin{abstract}
Detecting reclaimed slurs represents a fundamental challenge for hate speech detection systems, as the same lexical items can function either as abusive expressions or as in-group affirmations depending on social identity and context. In this work, we address Subtask B of the MultiPRIDE shared task at EVALITA 2026 by proposing a hierarchical approach to modeling the slur reclamation process. Our core assumption is that members of the LGBTQ+ community are more likely, on average, to employ certain slurs in a reclamatory manner. Based on this hypothesis, we decompose the task into two stages. First, using a weakly supervised LLM-based annotation, we assign fuzzy labels to users indicating the likelihood of belonging to the LGBTQ+ community, inferred from the tweet and the user bio. These soft labels are then used to train a BERT-like model to predict community membership, encouraging the model to learn latent representations associated with LGBTQ+ identity. In the second stage, we integrate this latent space with a newly initialized model for the downstream slur reclamation detection task. The intuition is that the first model encodes user-oriented sociolinguistic signals, which are then fused with representations learned by a model pretrained for hate speech detection. Experimental results on Italian and Spanish show that our approach achieves performance statistically comparable to a strong BERT-based baseline, while providing a modular and extensible framework for incorporating sociolinguistic context into hate speech modeling. We argue that more fine-grained hierarchical modeling of user identity and discourse context may further improve the detection of reclaimed language. We release our code at \href{https://github.com/LucaTedeschini/multipride}{https://github.com/LucaTedeschini/multipride}.
\end{abstract}

\section{Introduction}
\label{sec:introduction}

The automatic detection of hate speech and abusive language remains a central challenge in natural language processing, particularly in social media contexts where meaning is highly context-dependent. A recurring difficulty arises from the use of slur terms, whose interpretation cannot be determined solely from surface lexical content. In many communities, historically derogatory expressions are re-appropriated and used in neutral, positive, or identity-affirming ways, a phenomenon known as \textit{semantic reclamation}~\cite{PrideandPrejudiced, doi:10.3138/jld-2025-0501}. As a result, identical lexical items may convey either abuse or solidarity depending on pragmatic, social, and author-level factors. 
\newline
\newline
\textbf{Warning:} \textit{This paper contains examples of explicitly offensive content.}
\newline
\newline
This ambiguity poses a fundamental limitation for text-only classification models. While modern pretrained language models are effective at capturing local linguistic regularities, they often struggle to infer speaker intent when the interpretation of a slur depends on extra-linguistic context. Prior sociolinguistic research highlights that reclaimed usage is strongly associated with in-group membership, self-identification, and shared social knowledge~\cite{BIANCHI201435, cepollaro2023}.

Despite this, most computational approaches to hate speech detection operate under the assumption that such contextual information is either unavailable or implicitly encoded in the text itself. The MultiPRIDE shared task at EVALITA 2026~\cite{multipride,evalita2026overview} explicitly addresses this limitation by providing user biographies alongside tweets for Subtask B, enabling models to incorporate author-level context when determining whether a slur is used in a reclamatory or offensive manner.

In this work, we investigate whether decomposing the classification process into two hierarchical levels, assuming a directional dependency between user identity and language use, can improve performance over standard fine-tuning of a BERT-like model.

Our approach is motivated by the hypothesis that there exists a strong association between user identity and the reclamatory use of slurs. In particular, users who self-identify as part of, or closely aligned with, the LGBTQ+ community may be more likely to employ certain slurs in a reclamatory manner compared to users who are not. We further hypothesize that inferring such identity-related signals is comparatively easier when leveraging user biographies, where individuals often explicitly express aspects of their identity through self-descriptions, pronouns, symbols, or community-related markers.

Based on these assumptions, we adopt a hierarchical modeling strategy. First, we train a model to encode users into a latent space capturing signals related to LGBTQ+ self-identification. We then integrate this latent representation with a second model initialized with weights pretrained for hate speech detection. The underlying intuition is that the two models encode complementary perspectives on the input, one emphasizing sociolinguistic user context and the other focusing on abusive language patterns, whose fusion yields a richer representation for downstream slur reclamation detection.

Experiments on Italian and Spanish show that the proposed architecture achieves performance statistically equivalent to a strong BERT-based baseline. While this does not result in an improvement in aggregate evaluation metrics, it establishes a modular framework for incorporating user-level sociolinguistic context into hate speech modeling. We argue that extending this hierarchical approach to incorporate additional contextual signals, such as multimodal content or social network information, may further enhance the detection of reclaimed language in future work.

\section{Task and Data}
\subsection{Task Definition}
The shared task addresses slur reclamation detection in social media. Given a tweet containing at least one LGBTQ+ related term, the system must predict whether the term is used in a reclamatory manner (i.e., in-group, non-derogatory, or empowering usage) or in a non-reclamatory manner (e.g., derogatory or otherwise non-reclaiming uses).

The task is divided into two subtasks: Task A, which relies solely on tweet text, and Task B, which additionally incorporates contextual information such as user biographies. Building on the motivation outlined in Section~\ref{sec:introduction}, we restrict the scope of our contribution to Task~B.

\subsection{Data Characteristics}
The dataset provided for the challenge exhibits a substantial class imbalance, as summarized in Table~\ref{tab:eda}.

\begin{table}[h!]
\centering
\caption{Label distribution for the slur reclamation task. \emph{Non-reclamatory} instances correspond to uses of LGBTQ+ related terminology that are derogatory or not employed in a self-affirming or in-group manner.}
\label{tab:eda}
\begin{tabular}{lcc}
\toprule
\textbf{Language} & \textbf{Non-reclamatory} & \textbf{Reclamatory} \\
\midrule
Italian  & 879 (80.95\%) & 207 (19.05\%) \\
Spanish  & 743 (84.83\%) & 133 (15.17\%) \\
\bottomrule
\end{tabular}
\end{table}

For both languages, over 80\% of the tweets contain terms used in a non-reclamatory manner. This imbalance necessitates the adoption of mitigation strategies, as the objective is to minimize both false positives (incorrectly flagging reclaimed in-group usage) and false negatives (failing to detect harmful or derogatory uses). Table~\ref{tab:examples_with_bio} presents representative examples from the dataset in both Italian and Spanish. 

To create training, validation, and test sets, we performed two consecutive stratified splits on the original dataset resulting in an overall 70\% / 15\% / 15\% split for train, validation, and test, respectively. 
Stratification ensures that each subset maintains the same label distribution as the original dataset, mitigating issues due to class imbalance reported in Table~1. 
For all experiments, the train/validation/test splits were fixed using a set seed to ensure reproducibility.

\begin{table*}[t]
\centering
\caption{Representative examples of tweets with their labels and corresponding user biographies. Tweets and biographies are truncated for readability.}
\label{tab:examples_with_bio}
\small 
\begin{tabular}{l c p{5cm} p{5cm}}
\toprule
\textbf{Language} & \textbf{Label} & \textbf{Tweet} & \textbf{Biography} \\
\midrule
Italian & Non-reclamatory & ORA, I CLANDESTINI, AVRANNO SUBITO DIRITTI DI CITTADINANZA ITALIANA [...] & Voglio un governo che tuteli noi italiani \\ 
Italian & Reclamatory & Io amo essere frocio, viva i finocchi e tutt* i/le/l\textschwa\ mi\textschwa\ amic\textschwa\ ricchion\textschwa & Cogito ergo cum -- he/him \\ 
Spanish & Non-reclamatory & \#orgullogay no me jodan puedo sentir orgullo de ser un buen padre [...] & SACERDOTE MAYOR DE LA RELIGIÓN YORUBA [...] \\ 
Spanish & Reclamatory & Hoy amanecí más marica que todos los días! [...] & ARQUITECTO -- Popayán, Colombia. \\ 
\bottomrule
\end{tabular}
\end{table*}

\subsection{Why User Content Matters}
A qualitative inspection of the data suggests that lexical cues alone are often insufficient to resolve reclamation. Tweets are frequently short, elliptical, ironic, or contextually ambiguous, and may include quoted speech or re-contextualized language. In such settings, user self-descriptions provide additional context that can be indicative of stance and community alignment.

Profile text may contain self-disclosed signals, such as pronouns, identity-related descriptors, or community markers, that correlate with in-group usage patterns. In this study, however, we conceptualize these signals strictly as contextual cues derived from text, rather than as veridical or ground-truth indicators of personal identity.

\subsection{LLM-Assisted Proxy Labeling for the User Encoder}
Building on this intuition, we define an auxiliary proxy variable, referred to as an \emph{LGBTQ+ affiliation signal}, which is used exclusively to train the user encoder. Concretely, we automatically assign a binary label based on the concatenation of a user’s tweet and biography using an instruction-tuned Large Language Model (LLM), specifically DeepSeek-V3.2~\cite{deepseekai2025deepseekv32}. The two segments are concatenated using the separator token \texttt{[SEP]} between the tweet and biography. No additional separator or custom template is employed. The resulting label is intended to capture whether the available textual evidence suggests in-group affiliation or solidarity, rather than to infer or validate a protected attribute.

The exact prompts used for LLM-based annotation are provided in Appendix~\ref{sec:prompt}. 
Italian tweets were annotated using the original Italian prompt, while Spanish tweets were annotated using a Spanish translation of the prompt. 
For transparency and reproducibility, the Appendix includes both versions, with the Spanish prompt directly translated from the Italian text.

We explicitly acknowledge that automatic user profiling raises ethical and methodological concerns, including issues of privacy, consent, and the risk of stereotyping or misclassification. To mitigate these risks, we employ these labels solely as a form of weak supervision for representation learning, not as a user-level prediction target. The objective is to encourage the model to encode latent social and contextual signals useful for reclamation detection, without claiming accurate identification of individuals or verification of protected attributes. Accordingly, we do not report user-level outcomes and focus exclusively on aggregate task performance.

The resulting auxiliary label distribution is reported in Table~\ref{tab:lgbtq_belonging_distribution}.
\begin{table}[h!]
\centering
\caption{Distribution of proxy ``LGBTQ+ affiliation'' labels across languages. Labels are derived via LLM-assisted weak supervision and used to train the user encoder.}
\label{tab:lgbtq_belonging_distribution}
\begin{tabular}{lcc}
\toprule
\textbf{Language} & \textbf{Affiliated} & \textbf{Not Affiliated} \\
\midrule
Italian & 430 (39.60\%) & 656 (60.40\%) \\
Spanish & 362 (41.32\%) & 514 (58.68\%) \\
\bottomrule
\end{tabular}
\end{table}

Notably, this auxiliary distribution is more balanced than the primary task labels. At the same time, the auxiliary task is only imperfectly aligned with reclamation: while community alignment may increase the likelihood of reclaimed usage, it does not entail it (e.g., in cases of neutral self-reference, descriptive mentions, or internal criticism). This imperfect alignment is beneficial in practice, as it encourages the model to encode broader contextual information rather than collapsing to a narrow lexical heuristic.

\section{System Description}
We formulate slur reclamation detection as a binary classification task whose interpretation depends on both (i) the local linguistic context of the tweet and (ii) author-level contextual cues derived from user self-descriptions. To model these signals, we adopt a dual-encoder architecture with gated representation-level fusion.

\subsection{Baseline Model}
\label{cap:baseline}
As a reference system, we fine-tune a BERT-like encoder using a class-weighted loss function, following the recommendations of the MultiPRIDE shared task organizers. Italian and Spanish are treated independently, using language-specific pretrained models: \textit{nickprock/setfit-italian-hate-speech}~\cite{tunstall2022efficientfewshotlearningprompts} for Italian and \textit{cardiffnlp/twitter-xlm-roberta-base}~\cite{barbieri-etal-2022-xlm} for Spanish.

Each baseline model is trained using identical hyperparameters. To reduce variance due to random initialization, we report results averaged over five runs with different random seeds. The input to the model consists of the concatenation of the tweet text and the corresponding user biography. Performance results are reported in Table~\ref{tab:baseline_performances}.

\begin{table}[h]
\centering
\caption{Performance of the baseline BERT-based models on the development set, averaged over five random seeds (mean $\pm$ standard deviation).}
\label{tab:baseline_performances}
\begin{tabular}{lcc}
\toprule
\textbf{Language} & \textbf{Accuracy} & \textbf{F1 score} \\
\midrule
Spanish & 0.835 $\pm$ 0.034 & \textbf{0.671 $\pm$ 0.033} \\
Italian & 0.942 $\pm$ 0.017 & \textbf{0.904 $\pm$ 0.027} \\
\bottomrule
\end{tabular}
\end{table}

A noticeable performance gap emerges between the two languages, with the Italian baseline already achieving relatively strong scores. This discrepancy may stem from differences in annotation practices, data distributions, model pretraining domains, or language-specific sociolinguistic factors. Overall, the strength of the baseline makes further improvements non-trivial and motivates architectural modifications rather than incremental hyperparameter tuning.

\subsection{Dual-Encoder Architecture}
\begin{figure*}[!htbp]
    \centering
    \includegraphics[width=1.0\textwidth]{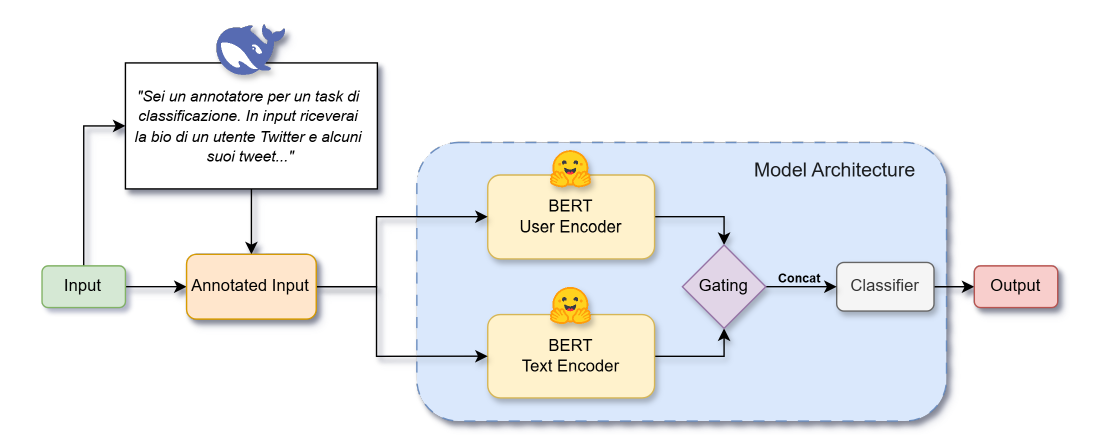}
    \caption{Overview of the proposed dual-encoder architecture. A text encoder and a user encoder process the concatenation of tweet and user biography in parallel. Their final hidden representations are fused through a learned gating mechanism to produce a user-aware representation for slur reclamation classification.}
    \label{fig:dual_encoder_architecture}
\end{figure*}

The proposed architecture consists of two parallel BERT-like encoders that process identical inputs, the concatenation of the tweet and the user biography, but are optimized under different training signals. Separate dual-encoder models are constructed for Italian and Spanish, reusing the same language-specific backbones adopted in the baseline.

The key intuition is that the two encoders learn complementary representations: one encoder is optimized directly for slur reclamation detection, while the other is encouraged to encode broader user-related sociolinguistic signals via an auxiliary proxy task. These representations are subsequently combined through a learned gating mechanism.

Although the user encoder also processes the tweet in addition to the biography, this architectural choice was informed by preliminary experiments indicating degraded performance when relying exclusively on the biography. Incorporating both segments enables the user encoder to leverage the full range of contextual signals (including lexical style, syntactic structure, and phrasing patterns) that underlie the LLM-derived proxy labels. Consequently, the text encoder remains dedicated to the primary task, thereby yielding complementary and mutually informative representations.

\subsubsection{User Encoder}
To induce user-oriented representations, we fine-tune the user encoder using the auxiliary proxy labels previously described. These labels, derived via LLM-assisted weak supervision, reflect whether the combined tweet and biography text suggests in-group affiliation or solidarity, without asserting ground-truth identity.

Performance on the proxy task is reported in Table~\ref{tab:user_encoder_pretrain_performances}.

\begin{table}[h]
\centering
\caption{Performance of the user encoder on the proxy ``LGBTQ+ affiliation'' task, averaged over five random seeds (mean $\pm$ standard deviation).}
\label{tab:user_encoder_pretrain_performances}
\begin{tabular}{lcc}
\toprule
\textbf{Language} & \textbf{Accuracy} & \textbf{F1 score} \\
\midrule
Spanish & 0.792 $\pm$ 0.038 & \textbf{0.784 $\pm$ 0.041} \\
Italian & 0.753 $\pm$ 0.021 & \textbf{0.736 $\pm$ 0.024} \\
\bottomrule
\end{tabular}
\end{table}

Although this auxiliary task is imperfectly aligned with slur reclamation, the encoder learns latent representations that capture broader contextual and sociolinguistic cues. When integrated into the dual-encoder architecture, these representations provide complementary information to the text-focused encoder.

We initially experimented with freezing the user encoder during downstream training, under the assumption that preserving its learned representations would prevent task-specific overfitting. However, empirical results consistently showed that allowing the user encoder to be fine-tuned jointly with the rest of the model yielded better performance. Consequently, all final experiments employ fully trainable encoders.

\subsection{Training Recipe}

\subsubsection{Hyperparameters}
Both encoders are trained using standard fine-tuning settings. We use a learning rate of $2 \times 10^{-5}$, batch size of $8$, and weight decay of $0.1$. For the dual-encoder model, we adopt a Linear Probing–Fine-Tuning (LPFT) strategy: the fusion and classification layers are first trained with frozen encoders, followed by joint fine-tuning of the entire architecture using a reduced learning rate of $5 \times 10^{-6}$.

\subsubsection{Loss Function}
To address class imbalance, we experimented with Focal Loss~\cite{lin2018focallossdenseobject} as an alternative to standard Cross Entropy. For Spanish, Cross Entropy achieved higher F1 scores than Focal Loss, while for Italian the two losses yielded comparable performance. Given the lack of consistent improvements, Cross Entropy was adopted for all final submissions.

\subsubsection{Feature Fusion via Learned Gating}
The representations produced by the two encoders are combined using a learned gating mechanism. Let $\mathbf{h}_{\text{text}}$ and $\mathbf{h}_{\text{user}}$ denote the final hidden states of the text and user encoders, respectively. The gating vector $\mathbf{g}$ is computed as:

\begin{equation}
    \mathbf{g} = \sigma\left(\mathbf{W}_2 \cdot \tanh\left(\mathbf{W}_1 \cdot [\mathbf{h}_{\text{text}} :: \mathbf{h}_{\text{user}}]\right)\right),
\end{equation}

where $[\cdot :: \cdot]$ denotes concatenation. The fused representation is then obtained as:

\begin{equation}
    \mathbf{h}_{\text{fused}} = \mathbf{g} \odot \mathbf{h}_{\text{text}} + (1 - \mathbf{g}) \odot \mathbf{h}_{\text{user}},
\end{equation}

with $\odot$ indicating element-wise multiplication. This mechanism allows the model to dynamically balance reliance on textual content and user-level contextual signals on a per-instance basis.

\subsubsection{Training Procedure}
The overall training pipeline consists of three stages: (i) fine-tuning a baseline text encoder on the slur reclamation task, (ii) training the user encoder on the auxiliary proxy task to induce user-oriented representations, and (iii) training the fused dual-encoder model on the primary reclamation task, jointly optimizing the gating mechanism, classifier, and both encoders.

\section{Results}
\label{sec:results}
To provide a comprehensive overview of model performance, we report results on the development set averaged over five runs for both the baseline and the proposed dual-encoder architecture in Table~\ref{tab:final_performances}. In addition, we report performance on the official test set released by the shared task organizers in Table~\ref{tab:real_test_performances}.

\begin{table}[ht]
\centering
\caption{Development set performance (macro F1) of baseline and dual-encoder models, averaged over five random seeds. $p$-values indicate statistical significance with respect to the baseline.}
\label{tab:final_performances}
\begin{tabular}{lcccc}
\toprule
\textbf{Model} & \textbf{Language} & \textbf{F1 (mean $\pm$ std)} & \textbf{$p$-value} \\
\midrule
Baseline     & Italian & 0.90 $\pm$ 0.03 & - \\
Baseline     & Spanish & 0.67 $\pm$ 0.04 & - \\
Dual Encoder & Italian & 0.88 $\pm$ 0.04 & 0.28 \\
Dual Encoder & Spanish & 0.64 $\pm$ 0.02 & 0.17 \\
\bottomrule
\end{tabular}
\end{table}

\begin{table}[ht]
\centering
\caption{Official test set performance (macro-averaged) of the dual-encoder model. $p$-values were not computed for this set.}
\label{tab:real_test_performances}
\begin{tabular}{lccc}
\toprule
\textbf{Language} & \textbf{Precision} & \textbf{Recall} & \textbf{F1-score} \\
\midrule
Italian & 0.88 & 0.83 & \textbf{0.85} \\
Spanish & 0.69 & 0.70 & \textbf{0.70} \\
\bottomrule
\end{tabular}
\end{table}

\subsection{Results Discussion}
The results indicate that the proposed dual-encoder architecture achieves performance statistically equivalent to the baseline on the development set ($p > 0.05$), which itself constitutes a strong benchmark for the task. On the official test set, the dual-encoder model maintains performance comparable to the baseline, confirming that introducing additional modeling complexity does not degrade results. Overall, this suggests that the dual-encoder framework successfully incorporates user-level context without harming predictive accuracy.

While the dual-encoder model does not yield improvements in aggregate evaluation metrics, it demonstrates that incorporating user-oriented representations via hierarchical and weakly supervised modeling can be achieved without sacrificing classification accuracy. We view this outcome as a validation of the architectural framework rather than as an endpoint. In particular, the modular nature of the approach allows for future extensions, such as alternative fusion mechanisms, more refined proxy tasks, or the integration of additional contextual encoders, that may more fully exploit user-level information.

\subsection{Differences Between Italian and Spanish}
\label{sec:diff_italian_spanish}
The substantial performance gap observed between the Italian and Spanish models is likely attributable to differences in the underlying baseline architectures and pretraining regimes rather than to intrinsic linguistic properties of the two languages.

The Italian baseline relies on a SetFit-based model, which is specifically designed to perform well in low-resource and few-shot settings. Moreover, this model has been previously fine-tuned on a hate speech detection task, which is closely related to slur reclamation detection. Such task-adjacent pretraining plausibly facilitates positive transfer, as both tasks require sensitivity to nuanced, context-dependent uses of potentially offensive language.

In contrast, the Spanish baseline is based on a general-purpose multilingual transformer that does not benefit from comparable task-specific pretraining and is less explicitly optimized for few-shot learning scenarios. As a result, the Spanish model may struggle to generalize effectively under limited supervision. We therefore attribute the observed performance disparity primarily to architectural and pretraining differences, rather than to language-specific sociolinguistic factors.

\subsection{Error Analysis}
We conclude by examining representative misclassified examples from the Italian dataset. Despite the strong overall performance, the model occasionally fails on instances that contain overtly derogatory slur usage and are unambiguously labeled as non-reclamatory.

For example, the following tweets were incorrectly classified as reclamatory despite exhibiting clear hostile intent:
\begin{quote}
\small
\textit{Caro \#Zan, quando froci, ricchioni, pederasti e lesbiche impareranno a rispettare i diritti altrui io smetterò di chiamarli così...} \\
\textit{Maledetti Grullini... un'accozzaglia arcobaleno di froci e lesbiche. Possa il Coronavirus fare pulizia tra di voi...} \\
\textit{Ronaldo è solo un travestito da gay pride che se la fa con quell'altro effeminato di Dybala}
\end{quote}

These errors suggest that the model may, in some cases, overestimate the likelihood of reclamatory usage when specific slur terms are present, independently of their surrounding discourse. One possible explanation is that certain lexical items are underrepresented in explicitly non-reclamatory contexts within the training data, leading to biased latent representations. Another contributing factor may be the influence of user-level contextual signals encoded by the user encoder, which could inadvertently dominate the gating mechanism even when the tweet content is clearly hostile.

While the precise cause of these errors cannot be conclusively identified within the scope of this work, they highlight an important limitation of representation-based fusion approaches: contextual signals, if not sufficiently constrained, may override strong local indicators of abuse. Future work could address this issue by introducing explicit regularization mechanisms, contrastive objectives, or token-level supervision to better disentangle reclaimed usage from overtly denigratory language.

\section{Conclusion}
In this work, we presented a dual-encoder architecture for slur reclamation detection that explicitly integrates textual content with user-level contextual information derived from profile descriptions. We evaluated the approach on the MultiPRIDE dataset for Italian and Spanish, showing that the proposed model achieves strong performance across both languages.

While the dual-encoder architecture does not yield statistically significant improvements over a standard BERT-based baseline (which itself constitutes a highly competitive benchmark) it demonstrates that user-oriented contextual representations can be incorporated without degrading performance. These findings suggest that hierarchical and representation-based modeling of sociolinguistic context is a viable direction for slur reclamation detection.

The primary contribution of this work lies in the architectural framework rather than in immediate performance gains. By explicitly separating text-based and user-oriented representations and combining them through a learned fusion mechanism, we provide a flexible and extensible structure for modeling context-dependent language phenomena.

\subsection{Future Work}
The strength of the baseline indicates that slur reclamation detection is already well addressed by current pretrained language models when sufficient textual context is available. Nevertheless, the modular design of the proposed dual-encoder architecture opens several avenues for future research.

Future work may focus on developing more expressive fusion strategies for heterogeneous latent spaces, as well as on refining the auxiliary supervision used to induce user-oriented representations. Additionally, the architecture naturally supports the integration of further contextual encoders, such as those operating on social network structures or multimodal inputs (e.g., images). We expect that more effective integration of these diverse information sources may enable models to capture finer-grained distinctions in reclaimed language use and ultimately surpass current strong baselines.

\section*{Declaration on Generative AI}
During the preparation of this work, the author(s) used Google Gemini 3.0 in order to: Grammar and spelling check. After using these tool(s)/service(s), the author(s) reviewed and edited the content as needed and take(s) full responsibility for the publication’s content.

\begin{acknowledgments}
This work was supported by the project ``IPCEI-CIS - Progetto Villanova'' (Beneficiary: Tiscali Italia S.p.A., Prog. n. SA. 102519 -- CUP B29J24000850005).
\end{acknowledgments}

\bibliography{sample-ceur}

\appendix
\section{LLM Annotation Prompts}
\label{sec:prompt}

\subsection*{Italian prompt}
\begin{tcolorbox}[colback=gray!5, colframe=black, boxrule=0.4pt]
\ttfamily\small
Sei un annotatore per un task di classificazione.\\
In input riceverai la bio di un utente Twitter e alcuni suoi tweet\\
Il tuo compito è decidere se l'utente in questione fa parte o meno della comunità LGBT.\\
Basa la tua scelta sul modo di scrivere, sul contenuto della bio e dei tweet, e su tutti i fattori che ritieni rilevanti.\\
L'input che ti fornirò sarà nel formato TWEET - BIO.\\

L'output che voglio è semplicemente un numero:\\
0 se l'utente NON appartiene alla comunità LGBT,\\
1 se invece appartiene alla comunità.\\

Esempio di interazione:\\
INPUT: "fuck gender rules and the rules of society || bts || exo" -
       "pansexual, genderqueer and polyamorous [LGBTQ+ flag] || she/her || unito dams"\\
OUTPUT: 1
\end{tcolorbox}

\subsection*{Spanish prompt}
\begin{tcolorbox}[colback=gray!5, colframe=black, boxrule=0.4pt]
\ttfamily\small
Eres un anotador para una tarea de clasificación.\\
En la entrada recibirás la biografía de un usuario de Twitter y algunos de sus tuits.\\
Tu tarea es decidir si el usuario en cuestión forma parte o no de la comunidad LGBT.\\
Basarás tu decisión en la forma de escribir, el contenido de la biografía y de los tuits,
y en todos los factores que consideres relevantes.\\
La entrada que te proporcionaré tendrá el formato TWEET - BIO.\\

La salida que quiero es simplemente un número:\\
0 si el usuario NO pertenece a la comunidad LGBT,\\
1 si sí pertenece a la comunidad.\\

Ejemplo de interacción:\\
ENTRADA: "fuck gender rules and the rules of society || bts || exo" -
         "pansexual, genderqueer and polyamorous [LGBTQ+ flag] || she/her || unito dams"\\
SALIDA: 1
\end{tcolorbox}

\paragraph{Note.}
The token \texttt{[LGBTQ+ flag]} is a placeholder for the LGBTQ+ flag emoji,
encoded by the Unicode sequence
\texttt{U+1F3F3 U+FE0F U+200D U+1F308}.

\end{document}